%% file: main.tex
\documentclass[a4paper]{article}
\usepackage{INTERSPEECH2021}

\usepackage{microtype}
\usepackage{graphicx}
\usepackage{xcolor}
\usepackage{hyperref}
\usepackage{makecell}
\usepackage{rotating}
\usepackage{colortbl}
\usepackage{tablefootnote}
\usepackage{xstring}
\usepackage{latexsym}
\usepackage{ragged2e}
\usepackage{linguex}
\usepackage{arabtex}
\usepackage{utf8}
\usepackage[utf8]{inputenc}
\setcode{utf8}

\usepackage{enumitem}

\setlist[itemize]{align=parleft,left=0pt..1em}
\usepackage{multirow}

\newcommand{\arastr}[1]{{\small \<#1>}}
\title{Towards One Model to Rule All: \\Multilingual Strategy for 
Dialectal Code-Switching Arabic ASR}

\name{Shammur Absar Chowdhury$^{1}$, Amir Hussein$^{1,2}$, Ahmed Abdelali$^{1}$, Ahmed Ali$^{1}$ }
\address{
  $^1$Qatar Computing Research Institute, Qatar \\
  $^2$KanarI AI , California}
\email{\{shchowdhury, aabdelali, amali\}@hbku.edu.qa, amir@kanari.ai}

\begin{document}
\maketitle


%
\begin{abstract}

With the advent of globalization, there is an increasing demand for multilingual automatic speech recognition (ASR), handling language and dialectal variation of spoken content. Recent studies show its efficacy over monolingual systems. In this study, we design a large multilingual end-to-end ASR using self-attention based conformer architecture. We trained the system using Arabic (Ar), English (En) and French (Fr) languages.
We evaluate the system performance handling: (i) monolingual (Ar, En and Fr); (ii) multi-dialectal (Modern Standard Arabic, along with dialectal variation such as Egyptian and Moroccan); (iii) code-switching -- cross-lingual (Ar-En/Fr) and dialectal (MSA-Egyptian dialect) test cases, and compare with current state-of-the-art systems. Furthermore, we investigate the influence of different embedding/character representations including character {\em vs} word-piece; shared {\em vs} distinct input symbol per language.
Our findings demonstrate the strength of such a model by outperforming state-of-the-art monolingual dialectal Arabic and code-switching Arabic ASR.

\end{abstract}

\noindent\textbf{Index Terms}: multilingual, multi-dialectal, code-switching, conformer, E2E, speech recognition

\input{sections/introduction}

\input{sections/methodology}
\input{sections/data}

\input{sections/experiments}
\input{sections/results}
\input{sections/conclusion}

\newpage


\bibliographystyle{IEEEtran}

\bibliography{mybib}

\end{document}

%% file: sections/introduction.tex
\section{Introduction}

Multilingual ASR \cite{datta2020language, pratap2020massively,li2019bytes,kannan2019large,toshniwal2018multilingual,cho2018multilingual,bourlard2011current,heigold2013multilingual,lin2009study,burget2010multilingual} have shown remarkable improvement over monolingual system and has recently gained immense popularity.
A single model capable of learning from many languages has been a motivation for multi- and cross-lingual speech communities for many decades. With the recent success of end-to-end models over the hybrid systems in monolingual settings \cite{heigold2013multilingual}, along with the availability of large multilingual speech datasets, more focus has shifted towards leveraging multiple languages for a more robust, language-agnostic all-in-one system. 

Studies like \cite{toshniwal2018multilingual,kannan2019large} used seq2seq and RNN-Transducer \cite{graves2012sequence} models for Indian languages, and used language id as an additional input to improve the performance. A recent study \cite{pratap2020massively}, designed a multilingual system covering 51 languages and suggested that such a system can benefit the performance of low-resourced languages. 

However, training such a model to cover all the languages ($\approx 7K$), is computationally challenging, and is hard to maintain. Furthermore, having a single model supporting all languages may not be a practical need for regional ASR -- e.g., Indic and Arabic languages -- where the language with-in is very rich in dialects and also has been heavily influenced by a handful of languages due to globalization and/or colonization. 

In such multilingual (and multidialectal) communities, code-switching -- a speaker switches from one language (dialects) to another within an utterance (intrasentential) -- is also a very common phenomenon. Recently, in addition to multilingual ASR, attention has been paid to design code-switching (CS) ASR. Studies have been conducted for Mandarin-English \cite{li2013improved}, Hindi-English \cite{sreeram2020exploration}, and French-Arabic \cite{amazouz2017addressing} and little to no prior work in dialectal code-switching. 

One of the major challenges in designing CS ASR is the lack of CS training data. This drawback restricts the exploitation of End-to-End (E2E) systems. However, recent studies like \cite{sreeram2020exploration} models Hindi-English CS using E2E attention model, and \cite{chan2016listen} uses context-dependent target to word transduction, factorized language model and code-switching identification. The authors in \cite{zhou2020multi} proposed two symmetric language-specific encoders to capture the individual language attributes in a transformer-based architecture. 

Unlike the aforementioned studies, we investigate how to design a multilingual ASR and leverage it to enhance the performance of the ASR in dialectal and code-switching contents, without any external language identification input or code shift recognition. For this study, \textit{Arabic} is an appropriate language due to its uniqueness as a shared language with 22 countries and having more than 20 mutually incomprehensible dialectal variety with modern standard Arabic (MSA) being the only standardized dialect \cite{badawi2013modern}, thus posing a unique set of challenges \cite{ali2021connecting}. Furthermore, the dialects in North African regions (e.g. Morocco, Algeria) are heavily influenced and adapted to the French language. French and English also act as lingua-franca with non-Arabs or non-native dialectal speakers and thus commonly use in spoken contents like broadcast news or other spoken contents.

\noindent Therefore, in this study, we :
\setlength\itemsep{-0.0em}
\begin{enumerate}[label=(\alph*)]

\item Designed an \textit{end-to-end ASR} system supporting dialectal Arabic (\textbf{Ar}), English (\textbf{En}) and French (\textbf{Fr}) languages. We trained the model using a self-attention based conformer architecture and benchmark the system in: {\em(i)} Monolingual contents: Ar, En, and Fr only, {\em(ii)} Dialectal Arabic (DA) contents: Egyptian, Moroccan and MSA, {\em(iii)} Code-switching contents: cross-lingual (Ar$\Leftrightarrow $En/Fr) and in dialectal CS (MSA$\Leftrightarrow $Egyptian dialect).

\item Compared \textit{different character representation space} (e.g., shared {\em vs} distinct input symbol for Latin languages, character {\em vs} word-piece tokenization) to investigate its influence on model performance. 

\item Analyzed the effect of inconsistent ASR output -- resulting in the same word being transcribed using different writing systems, on the ASR performance measure. For this, in addition to the word error rate, WER, we benchmark the CS ASR results with transliterated WER (TW).

\end{enumerate}
This is the first study to benchmark the performance of a multilingual ASR for dialectal and code-switching Arabic test sets. The proposed model outperforms current Arabic state-of-the-art E2E ASR. Without any adaptation, the model gives a comparable performance in heavily dialectal datasets. Our finding also suggests, with little to no code-switching in training data, the multilingual strategy is capable of modeling both cross-lingual and dialectal code-switching without any language/dialect identification input. With this study, we release two new code-switching datasets. This study is the potential benchmark for future dialectal and code-switching Arabic ASR.

%% file: sections/methodology.tex
\section{End-to-End Acoustic Model}
\label{ssec:acmodel}
To train a dialectal code-switching Arabic ASR with multilingual support, we adopted an end-to-end convolution-augmented transformer (conformer)\footnote{We also explored transformer encoders and obtained a similar pattern with all test sets. However, as the conformer outperforms the transformer encoder, for brevity, we are reporting all the results with conformer ASR. } architecture, consisting of a number of conformer encoders and transformer decoders, as proposed in \cite{conformer}. In the architecture, given input feature, the decoder predicts the output $\widehat{Y}_t$, at $t$ time step conditioning on the final latent representation of the encoder and the previous output target sequence, i.e. $\widehat{Y}_{1\dots t-1}$, in an auto-regressive manner.

\noindent\textbf{Input:} From raw signals, we extract $83$-dimensional feature frames consisting of $80$-dimensional log Mel-spectrogram and pitch features \cite{ghahremani2014pitch} and apply cepstral mean and variance normalization (CMVN).
The acoustic features, $X$ is then transformed into sub-sampled sequence $\widehat{X} \in \mathbb{R}^{^{O_l \times D_i}}$ ($O_l$: length of the output sequence and $D_i$ is input feature dimension to the encoder) using the convolution sampling layer.

\noindent\textbf{Encoder-Decoder:} We then pass the features to the conformer encoders -- each comprise of a stack of four modules including a positionwise feed-forward (FFN) module, multihead self-attention (MHSA) module, a convolution operation (CONV) module, and another FFN module in the end. 
As for the decoder, we used transformers -- each with an extra masked self-attention layer in addition to a MHSA and a feed-forward layer.

\noindent\textbf{Training Loss:} To improve the system robustness, we train the system using a multi-task learning objective. We combined the decoder cross entropy (CE) loss $\mathcal{L}_{ce}=-log(P_{d}(Y|X))$ and the CTC loss \cite{graves2006connectionist} $\mathcal{L}_{ctc}=-log(P_{ctc}(Y|X))$, with weighting factor, $\alpha$, given the posterior probability of output sequence $Y$ of $X$ acoustic input: $\mathcal{L}=\alpha \mathcal{L}_{\mathrm{ctc}}+(1-\alpha) \mathcal{L}_{\mathrm{ce}}$

%% file: sections/data.tex
\section{Corpus}
\label{sec:dataset}

\subsection{Monolingual Datasets}
\noindent\textbf{English Data:} For training, validating and testing the ASR, we use TEDLIUM3\footnote{\url{https://openslr.magicdatatech.com/51/}}, LibriSpeech\footnote{\url{https://openslr.magicdatatech.com/12/}} (clean and other) and Multi-Genre Broadcast (MGB-1) \cite{bell2015mgb}
dataset. Details of train/dev/test split along with genre is presented in Table \ref{tab:data_description}.

\noindent\textbf{French Data:} To incorporate the French language, due to its influence in North African Arabic countries, we add a subset from CommonVoice-French\cite{ardila2019common} dataset to the train/dev set and also test it on their official test data. See details in Table \ref{tab:data_description}.

\noindent\textbf{MSA and Dialectal Arabic Data:} For the study, we use a subset of 642 hours of speech data, from a multi-genre broadcast dataset, QASR \cite{qasr} dataset of 2k hours, collected from Aljazeera Arabic news channel's archive, spanning over 11 years from 2004 until 2015. The data includes lightly supervised transcriptions, covering contents mostly in MSA ($\approx$ 70\%) and the rest in DA from regions like Egypt, Gulf, Levantine, and North Africa. 

\begin{table}[!hbt]
\centering
\caption{Data description for train, dev and test set. C: means CMI value. The duration: in speech hours. [*]  hours of data for multiple reference.
}
\vspace{-0.3cm}
\scalebox{0.7}{
\begin{tabular}{c|l||c|c|c}
\hline
\multicolumn{2}{c||}{Datasets} & Train (hrs) & Dev (hrs) & Test (hrs) \\ 
\hline
\multirow{1}{*}{\textbf{En}} & TEDL / LibriC/ LibriO/ MGB1 & 50/50/50/50 & 1/0.5/0.5/-- & 2.62/5.4/5.1/-- \\
\hline
\textbf{Fr} & CV & 100 & 2 & 20.64 \\ 
\hline
\multirow{4}{*}{\textbf{Ar}} & QASR & 642 & - & - \\
 & MGB2 & - & 4 & 9.57 \\
 & MGB3 & \begin{tabular}[c]{@{}c@{}}  4 [16.14]\end{tabular} & 2 & 5.78 \\ 
 & MGB5 & \begin{tabular}[c]{@{}c@{}}10.2 [38.47]\end{tabular} & 1.3 & 1.4 \\  
\hline
\multirow{3}{*}{\textbf{CS}} & ESCWA.CS & - & - & \begin{tabular}[c]{@{}c@{}}2.8 (C: 28)\end{tabular} \\
 & QASR.CS & - & - & \begin{tabular}[c]{@{}c@{}}5.9 (C: 30.5)\end{tabular} \\
 & DACS & - & - & \begin{tabular}[c]{@{}c@{}}1.5 (C: 36.5)\end{tabular} \\ 
\hline\hline
\multicolumn{2}{c|}{TOTAL} & 996.61 & 11.3 & 51.14 \\\hline
\end{tabular}
}
\label{tab:data_description}
\vspace{-0.3cm}
\end{table}

A small percentage of the dataset also contains intrasentential code-switching. We quantify the amount of code-switching present in the subset using corpus level \textit{Code-Mixing Index} (CMI), motivated by \cite{chowdhury2020effects,gamback2016comparing}. 
See in Table \ref{tab:data_description} for details.

Furthermore, we add two real ecological dialectal datasets, collected from YouTube, distributed over different genres\footnote{For Example: Cooking, Sports, Drama, TEDx talks among others.}:
\vspace{-0.2cm}
\setlength\itemsep{-0.3em}
\begin{itemize}
    \item Egyptian MGB3
    \cite{ali2017speech}: A $\approx 16$ hours dataset from 80 videos.
    \item Moroccan MGB5\cite{ali2019mgb}: A $\approx 13$ hours of speech from 93 YouTube videos. 
\end{itemize}


\subsection{Intrasentential Code-Switching}
\noindent\textbf{Between-Language Code Switching Data:} To evaluate the ASR performance in intrasentential code-switching instances, we test the ASR using two code-switching test sets:
\vspace{-0.2cm}
\begin{itemize}
    \item \textbf{ESCWA.CS}:\footnote{\label{note1} \url{https://arabicspeech.org/escwa}} $\approx 2.8$ hours of speech code switching corpus collected over two days of meetings of the United Nations Economic and Social Commission for West Asia (ESCWA) in 2019. The data includes intrasentential code alternation between Arabic and English. In cases of Algerian, Tunisian, and Moroccan native-speakers, the switch is between Arabic and French.  Our initial analysis shows that such phenomena is present in $\approx$35\% of the dialectal Arabic speech. Further investigation indicates that on average 22\% of the segments is English/French content 
    and other 78\% is dialectal Arabic. 
    The corpus level CMI of ESCWA.CS is 28. 
    \item \textbf{QASR.CS}:\footnote{\url{https://arabicspeech.org/qasr}} $\approx 5.9$ hours of code switching extracted from the Arabic broadcast news data (QASR) to test the system for code switching. An average of $30.5$ CMI-value is observed in the corpus level. The dataset also include some instances where the switch is between Arabic and French, however this type of instances are very rare occurrence. 
\end{itemize}

\noindent\textbf{Dialectal Code Switching Data:} 
Dialectal code-switching (DCS) is a common phenomena in Arab region. However, it is more challenging and under studied research.
Therefore to evaluate DCS from MSA to Egyptian dialect and vice versa, we tested the ASR with dialectal Arabic code-switching dataset (\textbf{DACS})\footnote{\url{https://github.com/qcri/Arabic_speech_code_switching}} \cite{chowdhury2020effects} of $\approx 1.5$ hours of speech. 
The CMI value for the overall corpus is 36.5. In addtion to DCS, the data also includes few instances of cross-lingual CS. These code alteration tokens are transliterated in Arabic in the distributed corpus.
For more details about the data mentioned in the section, refer tot Table \ref{tab:data_description}. 

%% file: sections/experiments.tex
\section{Experimental Setup}
\label{sec:experiments}

\subsection{Data Preparation}
For the acoustic modeling, we first augment the raw speech data with the speed perturbation, with speed factors of $0.9$, $1.0$, and $1.1$ \cite{ko2015audio}. We then use the augmented audio to extract the input features (log Mel-spectrogram with pitch) and again augmented with specaugment approach \cite{park2019specaugment}.
For transcription, we first cleaned the data (\textit{i}) removing all punctuation except the $\%$ and $@$ due to its verbatim usage, (\textit{ii}) removing diacritics (for Arabic), (\textit{iii}) transliterating all Arabic digits to Arabic numerals (e.g. \<١> to 1) and (\textit{iv}) converting all the Latin characters to lower case. 

\begin{figure}[hbt!]
\begin{center}
\vspace{-0.3cm}
\scalebox{0.35}{
\includegraphics[]{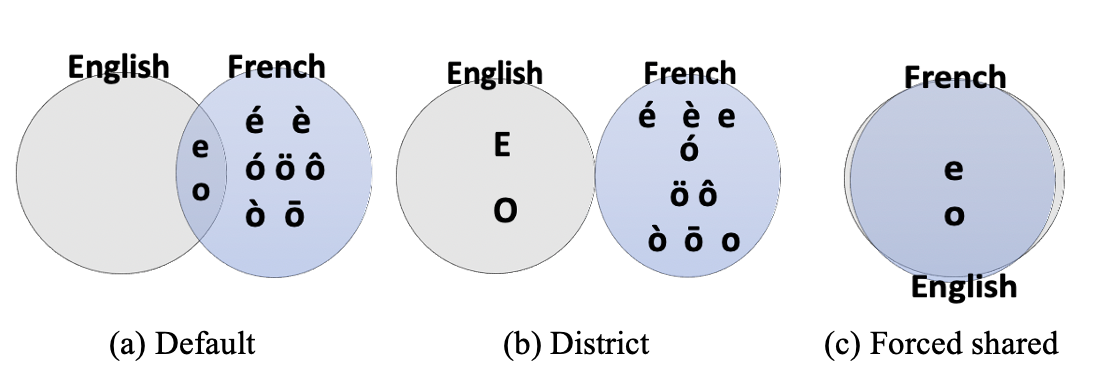}
}
\vspace{-0.4cm}
\caption{Illustration of the different character space.}

\label{fig:charspace} 
\end{center}
\vspace{-0.8cm}
\end{figure}

\subsection{Model Parameters}
\label{ssec:param}
We trained the end-to-end ASR using a Noam \cite{vaswani2017attention} optimizer for $50$ epochs with a learning rate of 5 with $20,000$ warmup steps and dropout-rate of 0.1. The trade-off weights, $\alpha$, for $\mathcal{L}$, we use a value of $\alpha=$0.3. The number of encoder, decoder layers and attention-heads differed based on the choice of large/small architecture.
As for the text tokenization, we used word-piece byte-pair-encoding (BPE)~\cite{kudo2018sentencepiece} for the multilingual ASR. 

\noindent\textbf{Large ASR Architecture:} For the large ASR (with $\approx 1,000$ hours of speech), we used 12 encoder layers and 6 decoder layers each with 2,048 encoder/decoder units from FFN and 8 attention heads with 512 transformation dimensions. For the architecture, we used 31 CNN kernals. For multilingual dialectal ASR, we opt for an BPE of $10K$.

\noindent\textbf{Small ASR Architecture:} For exploring the influence of different character space representation, we designed small-scale ASR using 8/4 encoder/decoder layers each with 2048 FFN units. As for the attention modules, we used 4 attention head with a dimension of 256. For the task, we opt for 15 CNN module kernals.

\subsection{Monolingual baseline models}
\label{ssec:monolingual}
To compare the performance of multilingual ASR with monolingual performance using the same English (200 hours) and French (100 hours) data, we trained individual monolingual ASR using the small conformer architecture described in  
Section \ref{ssec:param}. For the experiments we used a bpe size of 500, motivated by the ESPnet recipe for TEDLIUM data.
For the Arabic monolingual and dialectal models, we used reported study by \cite{hussein2021arabic}, using E2E transformer models trained with MGB2-train data.

\subsection{Shared/Distinct Character Space}
To explore the impact of using a shared or distinct character set, in the ASR performance, when dealing with multilingual and code-switching data, we experimented with different input character sets. We investigate three settings: (a) natural shared character (see Figure\ref{fig:charspace}(a) Default); and (b) distinct character (see Figure\ref{fig:charspace}(b) Distinct). Furthermore, we also compared with a strict variation of \textit{(a)}, where (c) all the nearby characters are mapped to one character representation (see Figure\ref{fig:charspace}(c) Forced shared). 

We randomly picked a subset of 100 hours of training speech, maintaining a similar distribution as the full dataset (i.e. 70\% Arabic, 20\% English, and the rest 10\% French, motivated by analysing ESCWA meeting recordings). We then trained ASR systems with each of the settings (in Figure\ref{fig:charspace}) and tested on the dev and test sets. 
We trained the model using the small conformer architecture and parameters (mentioned in Section \ref{ssec:param}) and a BPE size of 1000 ( i.e. $\frac{1}{10}$ of the large ASR).
In addition to BPE, we also use character-based encoding to compare the performance of default setting (Figure\ref{fig:charspace}a: Default), using a vocabulary size of 120 characters.

\subsection{Evaluation Measures}
We benchmarked the ASR using the conventional word error rate (WER). Furthermore, we also reported the transliteration WER (TW) for the code-switching test sets. We hypothesize that transliterating the En and Fr recognized tokens into Arabic, will help to disambiguate some code-switching errors introduced by the multilingual writing systems supported by the ASR.
For this, we created a simple Global Mapping File (GLM) to transliterate, using system proposed in \cite{dalvi-etal-2017-qcri} and the recognized outputs. For such baseline evaluation, we only considered English words and ignored any mixed-code tokens.


%% file: sections/results.tex
\begin{table}[hbt!]
\centering
\caption{Reported WER and TW on monolingual, dialectal Arabic datasets along with code-switching datasets. 
E2E:MGB2 represent transformer ASR trained with 1200hrs of MGB2. E2E:MGB2+TEDL: E2E:MGB2 + TEDLIUM3-train. $^{*}$: with added LSTM LM. Adapt: pretrained E2E:MGB2, adapted on the in-domain dialectal data.
}
\vspace{-0.3cm}
\scalebox{0.9}{
\begin{tabular}{l||c|c||c} 
\hline
\multicolumn{1}{c||}{\multirow{2}{*}{\textit{Tests}}} & \multicolumn{2}{c||}{Multilingual} & {SOTA} \\ 
\cline{2-4} 

\multicolumn{1}{c||}{} & \textbf{WER} & \textbf{TW} & \textbf{WER (\textit{Model})} \\ 
\hline\hline
MGB2 & \textbf{12.1} & -- & 12.5 (\textit{E2E:MGB2$^{*}$})\\ 

MGB3 & \textbf{35.2} & -- & 36 (\textit{Adapt}) \\ 

MGB5& 61.3 & -- & 57.2 (\textit{Adapt})\\ 
\hline\hline
QASR.CS& \textbf{26.4 }& \textit{25.8}  & -- \\ 
ESCWA.CS & \textbf{37.7} & \textit{37.3}  & 50.1 (\textit{E2E:MGB2+TEDL$^{*}$})\\ 
\multicolumn{1}{l||}{DACS} & \textbf{25.0} & \textbf{25.0} & 30.7 (\textit{E2E:MGB2})\\ 
\hline
\end{tabular}
}
\label{tab:asr_mlmd_results}
\vspace{-0.5cm}
\end{table}

\section{Results and Discussion}
\label{sec:results}
\subsection{ASR Performance}
The performance of the designed E2E multilingual ASR on dialectal Arabic and code-switching test sets are reported in Table \ref{tab:asr_mlmd_results}. The results are benchmarked using several state-of-the-art models discussed in \cite{hussein2021arabic}.

From WER, we observed that the proposed multilingual ASR outperforms the monolingual E2E transformer ASR (E2E:MGB2) -- trained with MGB2 ($2 \times$ more) data -- on the MGB2-test set by 0.4\% absolute WER. We also noticed an increase in performance (by 0.8\%) in MGB3-test compared to E2E ASR \cite{hussein2021arabic} adapted specifically for Egyptian dialect. The multilingual model gives a comparable performance to the adapted ASR (on MGB5 train), when tested on MGB5-test set.

One of the main drawbacks of a monolingual model is that it can not handle cross-lingual code-switching tasks. Therefore, we benchmarked the ASR on intrasentential code-switching tests using a bilingual transformer model \cite{ali2021csmodel}, trained using MGB2 and TEDLIUM3 train set. From the ESCWA.CS data, we see a significant decrease in WER when using the proposed ASR. This gain in performance reflects the capability of the multilingual model to handle both dialectal and code-switching data. 


As for the dialectal code-switching (DACS-test set), we noticed our multilingual ASR significantly outperforms a large monolingual E2E:MGB2 ASR by absolute 1.8\% WER. This again indicates the strength of the latent dialectal representation learned in the proposed ASR. 

In addition to the Arabic test sets, we evaluate the performance of the proposed ASR using monolingual En and Fr test sets, reported in Table \ref{tab:asr_mlmd_monolingual_results}. We compared the proposed multilingual ASR with its monolingual counterparts (see Section \ref{ssec:monolingual}) and also reported state-of-the-art results on these test sets.
We observed the multilingual ASR gives a comparable performance w.r.t. monolingual model, however, it is significantly accurate in spotting the language 
and no confusion was seen between different scripts for the En and Fr test sets.

\begin{table}
\centering
\caption{Reported WER for En and Fr testsets using monolingual ASR ; MLMD ASR; and the current SOTA (ESPnet) models using large conformer (LC) architecture. \* represents the WER obtained using transformer LM in addition to AM.  }
\vspace{-0.3cm}
\scalebox{0.9}{
\begin{tabular}{l||cc||c||c}
\hline
\multicolumn{1}{c||}{\multirow{2}{*}{\textit{WER}}} & \multicolumn{2}{c||}{Monolingual} & \multirow{2}{*}{\begin{tabular}[c]{@{}c@{}}Multilingual \\ (1khrs)\\Ar,En,Fr\end{tabular}} & \multirow{2}{*}{\begin{tabular}[c]{@{}c@{}}SOTA\\LC\end{tabular}} \\
\cline{2-3}
\multicolumn{1}{c||}{} & \multicolumn{1}{c}{\begin{tabular}[c]{@{}c@{}}En \\(200hrs)\end{tabular}} & \multicolumn{1}{c||}{\begin{tabular}[c]{@{}c@{}}Fr \\(100hrs)\end{tabular}} &  &  \\\hline\hline
TedL & 18.9 & -- & 19.8 & 7.6 \\
LibriC & 8 & -- & 8.6 & 2.1* \\
LibriO & 16.1 & -- & 16.5 & 4.9* \\
CV:Fr & -- & 16.2 & 18.2 & 14.8 \\\hline
\end{tabular}}
\label{tab:asr_mlmd_monolingual_results}
\vspace{-0.5cm}
\end{table}

\subsection{Error Analysis and Transliteration WER}
We studied different types of error in dialectal and code-switching test sets.
We noticed that in the dialectal dataset (mainly Moroccan: MGB5), most substitutions are due to inconsistencies between the dialectal and MSA orthographic and linguistic rules. We noticed substitutions in relative pronoun \arastr{اللي} (``Ally'' - ``which'') that does not exit in MSA and is replaced by its closest match \<لي> (``ly'' - ``for me''). We also noticed large numbers of substitutions due to the presence of vowels like Alif/Ya, such as the case of \arastr{نحنا} (``nHnA'' - ``we'') and  \arastr{نحن} (``nHn'' - ``we'') or\arastr{رح} (``rH'' - ``will'') and \arastr{راح} (``rAH'' - ``will'').
  Such errors are easily fixed with a GLM designed for Maghrebi dialects\footnote{We present WER after post-processing with such GLM, with a gain of $\approx 5\%$ in WER.}. 
It is commonly noted in dialectal Arabic that some substitutions are frequent in one or more dialects\cite{samih-etal-2017-neural} such as the case of ``\arastr{ث}'' or ``\arastr{ذ}'' that are substituted by ``\arastr{ت}'' or ``\arastr{س}'' and ``\arastr{ز}'' respectively. 
Another type of error, we noticed is partial/full transliteration of a word, e.g.
``Artificial'' to ``\arastr{ارت} ificial'' and ``Drones'' to ``\arastr{الدرونز}''. Such errors are noticed often and motivates the use of transliterated WER (TW). From the reported TW results in Table \ref{tab:asr_mlmd_results}, we noticed transliterating all outputs to one language smoothes out the rendering error and offers a better perspective of the output. However, this still does not handle partial transliterated ASR output (e.g., ``\arastr{ارت} ificial''). Thus indicating a need for a better CS evaluation metric.


\subsection{Effect of Different Character Representation Space}
\begin{figure}[hbt!]
\begin{center}
\scalebox{0.33}{

\includegraphics[]{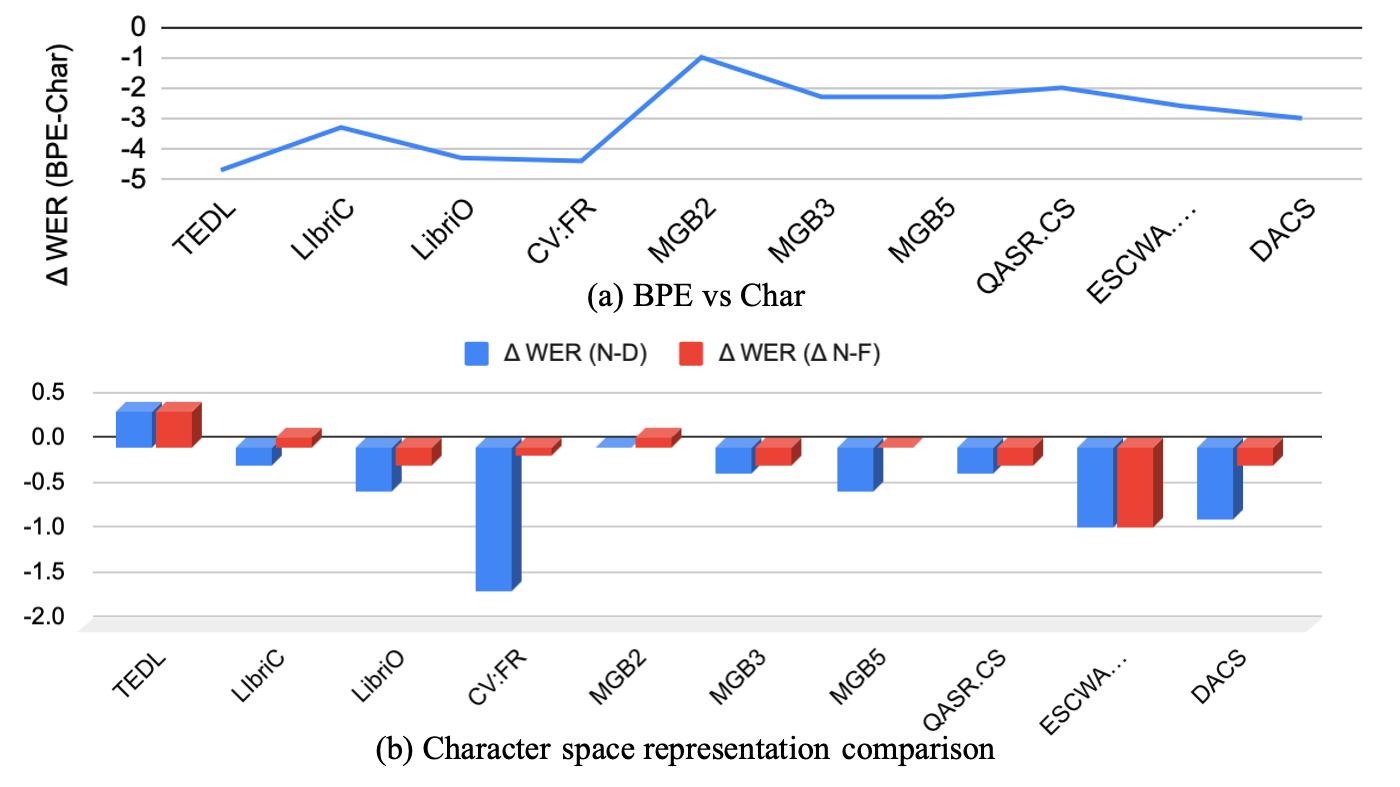}
}
\vspace{-0.4cm}
\caption{Change in WER $\Delta = WER(N) - WER(*)$ Figure \ref{fig:charspace_rslt} (a): BPE {\em vs} Char;  Figure \ref{fig:charspace_rslt}(b): naturally shared character (N) {\em vs} distinct (D) and forced merged (F) character set.}

\label{fig:charspace_rslt} 
\end{center}
\vspace{-0.8cm}
\end{figure}

We examined the effect of different strategies representing the Latin character space, and different tokenization techniques (BPE {\em vs} Char) (see Figure \ref{fig:charspace_rslt}a-b). We noticed using distinct character set (D) (Figure \ref{fig:charspace}b), WER increases significantly ($\Delta WER(N-D)$ is negative, meaning WER(D)$>$WER(N)) in the monolingual Fr. A similar observation is seen with the code-switching data, specially ESCWA.CS. We hypothesize such a change in performance is due to the presence of Fr CS and North African regional dialectal instances in ESCWA.CS data and the same is seen in MGB5 data which is also heavily influenced by Fr. 
Moreover, our result suggests that using the natural distinction (default settings in character space), the model is able to capture language-agnostic, yet discriminating representations.

In addition to character representation, we also noticed BPE is better in representing multilingual embeddings than character-based tokenization. Thus, validating our choice for using BPE as the tokenizer of the proposed ASR.

\vspace{-0.4cm}

\subsection{Key Observations }
Using the multilingual training strategy, we observed improved performance when tested on Modern Standard and Dialectal Arabic test sets. These findings indicate the strength of shared latent (dialectal) language representation in such an ASR. The WER in monolingual English and French along with intrasentential code-switching settings shows the proposed model is able to distinguish the input language with more success and has the potential to be the one system to replace them (monolingual, CS ASR) all. 
The advantage of using a shared language representation for modeling is also reflected from our study, using different character representation scripts. Our evaluation, using TW, also points out the drawbacks of using WER, as a measure for CS, in offering a better understanding of the model. Thus, signaling a need for better evaluation metrics for code-switching.

%% file: sections/conclusion.tex
\section{Conclusions}
\label{conclusion}

In this paper, we presented the first comprehensive study comparing multilingual ASR strategy to develop an E2E Arabic dialectal and code-switching ASR.
The study benchmarked the performance of the proposed multilingual ASR for dialectal and code-switching Arabic test sets. The multilingual E2E conformer model outperforms the current Arabic state-of-the-art E2E transformer ASR. 
Without any fine-tuning/adaptation, the model gives a comparable performance in dialectal datasets and is fully capable of handling dialectal and cross-lingual code-switching instances without the need for any/large training data.

Moreover, we explored the strength of such multilingual ASR in learning better latent representation -- gained from the multiple languages and its shared character space. With this study, we also release two new cross-lingual code-switching datasets. This dataset has the potential to benchmark future dialectal code-switching Arabic ASR.